\documentclass[10pt, a4paper]{article}

\usepackage{graphicx}
\usepackage{booktabs}
\usepackage{multirow}
\usepackage{adjustbox}
\usepackage{subfigure}
\usepackage[dvipsnames]{xcolor}

\usepackage[final]{lrec-coling2024} 

\title{Pitfalls of Conversational LLMs on News Debiasing}

\name{Ipek Baris Schlicht$^{1,2}$, Defne Altiok$^{1}$, Maryanne Taouk$^{3}$, Lucie Flek$^{4,5,6}$}
\address{$^1$Deutsche Welle, Bonn/Berlin, Germany \\
$^2$Universitat Politecnica de Valencia, Spain \\
$^3$ABC News, Australia \\
$^4$Conversational AI and Social Analytics (CAISA) Lab \\
$^5$Bonn-Aachen International Center for Information Technology (b-it), University of Bonn \\
$^6$The Lamarr Institute for Machine Learning and Artificial Intelligence}

\abstract{
This paper addresses debiasing in news editing and evaluates the effectiveness of conversational Large Language Models in this task. We designed an evaluation checklist tailored to news editors' perspectives, obtained generated texts from three popular conversational models using a subset of a publicly available dataset in media bias, and evaluated the texts according to the designed checklist. Furthermore, we examined the models as evaluator for checking the quality of debiased model outputs. Our findings indicate that none of the LLMs are perfect in debiasing. Notably, some models, including ChatGPT, introduced unnecessary changes that may impact the author's style and create misinformation. Lastly, we show that the models do not perform as proficiently as domain experts in evaluating the quality of debiased outputs.
 \\ \newline 
}

\begin{document}

\newif\ifproofread
\newcommand{\changemarker}[1]{%
\ifproofread
\textcolor{red}{#1}%
\else
#1%
\fi
}

\maketitleabstract
\proofreadfalse

\section{Introduction}
Biased news articles have the potential to significantly shape public opinion and discourse on various issues. Thus, professional news editors identify bias text spans in news articles before they are published. This task is particularly challenging, especially when editorial teams face constraints such as time-pressure and a lack of human resources.

Large Language Models (LLMs) have demonstrated outstanding performance even in the absence of labeled data, through zero-shot prompting. In many tasks, LLMs have surpassed the performance of the supervised models and have even employed as writing assistance~\cite{DBLP:journals/corr/abs-2208-01815,DBLP:journals/corr/abs-2305-13225}. In addition, conversational LLMs such as ChatGPT~\cite{ChatGPT} and GPT4~\cite{DBLP:journals/corr/abs-2303-08774} are user-friendly, making them accessible to non-technical experts like journalists who can use them without coding knowledge to aid in their tasks. As a result, many media companies have already begun experimenting ChatGPT for various journalistic tasks~\cite{polislse}. 
Limited studies have explored debiasing through text generation with conversational LLMs for the tasks such as hate speech~\cite{plaza-del-arco-etal-2023-respectful} and toxicity detection\cite{morabito-etal-2023-debiasing}. These studies explored zero-shot prompting with conversational LLMs. To the best of our knowledge, conversational LLMs have not been explored for news debiasing.

\begin{figure} [!htp]
    \centering
        \footnotesize
    \begin{tabular}{p{0.95\linewidth}}
\toprule
President Donald Trump gave states and local governments the right to reject refugees, \textcolor{blue}{but instead of saying no, most state and local officials have blindsided the administration by opting in, according to two former officials familiar with the matter.}
\\
\midrule
President Donald Trump allowed states and local governments the option to refuse refugees. \textcolor{red}{However, according to two former officials familiar with the matter, most state and local officials have chosen to accept refugees.}
\\
\bottomrule
    \end{tabular}
    \caption{Biased text \changemarker{where the usage ``blind-sided'' introduces bias by conveying a strong negative opinion about the actions of state and local officials} and its GPT4 debiased version which doesn't contain toxicity according to Perspective API. Debiasing changed the facts and the context  (factually incorrect statement highlighted in \textcolor{red}{red}, original version in \textcolor{blue}{blue}).}
    \label{tab:task_intro}
\end{figure}

\begin{table*}[!ht]
\footnotesize
    \centering
    \adjustbox{width=\textwidth}{
    \begin{tabular}{lll}
    \toprule
    \textbf{ID} & \textbf{Concept} & \textbf{Question} \\
    \midrule
    \multirow{ 4}{*}{\textbf{C1}} & \multirow{ 4}{*}{\textbf{Correcting Bias}} & \textbf{Does the model produce unbiased text? Grade 1-3}\\
    & & \textit{The text is unbiased. (3)} \\
    & & \textit{The text is partially biased. (2)} \\
    & & \textit{The text is highly biased. (1)} \\
    \midrule
    \multirow{ 4}{*}{\textbf{C2}} & \multirow{ 4}{*}{\textbf{Preserving Information}} & \textbf{Does the model change textual facts? Grade 1-3}  \\
    && \textit{The text facts are still present. (3)} \\
    && \textit{Some facts are missing. (2)} \\
    && \textit{Facts are completely missing. (1)} \\  
    \midrule
    \multirow{ 4}{*}{\textbf{C3}} & \multirow{ 4}{*}{\textbf{Preserving Context}}  & 
    \textbf{Does the model change the meaning of text? Grade 1-3} \\
    & & \textit{The meaning of the text is preserved.(3)} \\
    & & \textit{The meaning of the text is partially preserved. (2)} \\
    & & \textit{The meaning of the text is completely changed.(1)} \\
    \midrule  
       \multirow{ 4}{*}{\textbf{C4}} & \multirow{ 4}{*}{\textbf{Preserving Language Fluency}}  & \textbf{Does the model produce grammatically correct text? Grade 1-3} \\
    & &\textit{The text is grammatically correct. (3)} \\
    & &\textit{The text has few grammar issues. (2) } \\
    & &\textit{The text has many grammar issues. (1) } \\  
    \midrule
    \multirow{ 4}{*}{\textbf{C5}} &\multirow{ 4}{*}{\textbf{Preserving Author's Style}} & \textbf{Does the model harm the author's creativity? Grade 1-3} \\
    && \textit{No, the model did all necessary changes without harming author creativity. (3) }\\
    && \textit{The model corrected some of the texts that might hurt the creativity. (2)} \\
    && \textit{The model did unnecessary changes, and changed the text style. (1)} \\
    \bottomrule
    \end{tabular}}
    \caption{News editorial criteria for checking quality of debiasing.}
    \label{tab:bias_criteria}
\end{table*}

Standard evaluation metrics~\cite{min-etal-2023-factscore} such as ROUGE require a reference text for measuring generated text quality and lacks explanatory evaluation. \citet{morabito-etal-2023-debiasing} established an evaluation protocol for automatically assessing LLMs' consistency in debiasing for toxicity detection by using Perspective API\cite{perspective} as the evaluator. However, this protocol is limited to bias reduction and may not be suitable for the news domain. \changemarker{In the context of news bias, bias encompasses both overt bias such derogatory terms within text and latent biases that shape the language and framing of news stories~\cite{DBLP:conf/acl/RecasensDJ13}.} As shown in Figure~\ref{tab:task_intro}, news texts deemed biased may not contain toxicity \changemarker{but wording/phrasing could introduce bias}. Hence, tools such as Perspective API could fail to quantify bias reduction. Furthermore, the debiased text might produce misinformation by changing context and factuality and altering the author's writing style. Therefore, there is a need for evaluation criteria discerned editorial perspectives.

To address these issues, we investigate the following research questions (RQs):
(1) How well do conversational LLMs perform debiasing in the context of the news domain according to editorial criteria?
(2) Can conversational LLMs also serve as an evaluation tool for assessing the editorial quality of debiased articles?

Given the need for a domain-specific evaluation to assess the quality of conversational LLMs in news debiasing, we propose a set of evaluation criteria tailored to news editors. Since there is no publicly available news dataset for debiasing, we obtained text generations on a subset of the publicly available bias classification dataset using three popular conversational LLMs and a fine-tuned T5~\cite{2020t5}. Expert news editors from international media organizations ranked the models' outputs based on the editorial criteria. Additionally, we compared model outputs with expert assessments when the models were used as evaluation tools to check the quality of debiasing. Our results showed that despite conversational LLMs' proficiency in bias reduction, they sometimes generate misinformation and alter writing styles. Moreover, they can not assess debiased outputs as the experts do~\footnote{The code and the data are at \url{https://bit.ly/3vGphbw}}.

\section{Related Works}
The studies on media bias have primarily focused on two aspects: identifying biased text spans~\cite{spinde-etal-2021-neural-media, hamborg2020media, lei-etal-2022-sentence} and detecting political bias in news articles~\cite{chen-etal-2020-analyzing} or media outlets~\cite{baly2020written}. Only a few studies proposed methods for mitigating bias through article generation using transformer models. Among these studies, the earliest work by \citet{pryzant2020automatically} used BERT to identify subjective content and update the hidden layers of the model to generate unbiased text from Wikipedia. \citet{DBLP:conf/naacl/LeeBYMF22} applied a summarization method on articles from various political leanings to neutralize news.

\citet{plaza-del-arco-etal-2023-respectful} and \citet{morabito-etal-2023-debiasing} explored the potential of zero-shot prompting with LLMs, respectively for hate speech detection and reducing toxicity in user comments. Additionally, ~\citet{morabito-etal-2023-debiasing} established an evaluation protocol for evaluating consistency of LLMs on debiasing in the context of toxicity detection. The authors used Perspective API as the evaluator tool which provides toxicity scores for comment moderation. However, the protocol is limited to only to bias reduction. Furthermore, is not applicable within the news domain as news articles may not exhibit a toxic tone, yet they can still contain biases favoring certain groups, which need to be addressed before publication. In our work, we design evaluation criteria taking into account journalistic perspective to measure quality of debiased sentences.

Recently, researchers have explored LLMs as evaluators for assessing the quality of text generation in various applications~\cite{DBLP:journals/corr/abs-2402-01383,min-etal-2023-factscore} as an alternative solution to costly expert assessments. Motivated by this, we evaluate the conversational LLMs models as evaluators for assessing the quality of debiased sentences based on the journalistic criteria and compare them with our expert evaluation.

\section{Methodology}
We investigated three conversational LLMs for debiasing news sentences and paragraphs. Given sentences or paragraphs containing bias types such as epistemological, framing and demographic bias~\cite{pryzant2020automatically, spinde-etal-2021-neural-media,DBLP:conf/acl/RecasensDJ13}, the goal of the task was to generate an unbiased version of those sentences. The outcome of the sentences should be unbiased but other criteria should also be considered as important for news editors, such as preserving factuality, news' message, and not harming the authors' creativity, along with grammar changes. 

\subsection{News Editorial Criteria\label{sec:eval}}
As prior evaluation metrics are limited to news debiasing, we propose news editorial criteria. \changemarker{The editorial criteria were created during the implementation of BiasBlocker, which is a prototype AI-based news editor.~\footnote{\url{https://bit.ly/4aJttWD}}}.

\changemarker{The BiasBlocker team comprises experienced news editors and technologists from Deutsche Welle, ABC News and ARIJ. Since bias is a broad concept, to establish a common ground on the bias definitions and the corrections, the editors in the team created a codebook on bias types~\footnote{\url{https://bit.ly/49qcnvZ}} and guidelines for debiasing based on the prior studies~\cite{pryzant2020automatically, spinde-etal-2021-neural-media,DBLP:conf/acl/RecasensDJ13} and UN Guidelines~\footnote{\url{https://bit.ly/3PRks67}}. Hence, the bias types we focus on are primarily framing, epistemological, and demographic bias.}

\changemarker{We applied a pilot study on bias correction by using ChatGPT with the editors~\footnote{\url{https://bit.ly/43OMCnQ}}. The editors spotted the issues and refined the expectations for the news editor.} As outlined in Table~\ref{tab:bias_criteria}, we distilled these expectations into five criteria for assessing the quality of models in the context of debiasing for news editing.

Essentially, the editors expected the model to effectively remove any text spans that introduce bias into the content. However, they also had the expectation that this must refrain from adding new facts or removing vital information, as this could produce misinformation. Furthermore, the model must ensure that the meaning of the text remains intact. The debiased text must also be grammatically correct. Lastly, especially for those articles of opinion pieces or analyses, the model must respect and preserve the author's writing style and creativity. Otherwise, the model could discourage less experienced authors and harm the communication of the news message.

\noindent
\textbf{Evaluation Dataset.} Wiki Neutrality Corpus (WNC)~\citeplanguageresource{pryzant2020automatically} is the only publicly available dataset that contains biased samples and their debiased version by Wikipedia editors. Given that our research objective was to assess the LLMs in correcting bias within texts authored by news authors, WNC samples were not suitable for the evaluations. Therefore, we preferred the BABE dataset~\cite{spinde-etal-2021-neural-media} as the test set. BABE consists of sentences from news articles published by US publishers with different political leanings. Experienced media experts annotated the dataset; the dataset samples were labeled as biased or unbiased. The authors of the dataset provided two subsets. We chose the one annotated with more experts and randomly selected 50 biased sentences from this subset for the evaluations.
\subsection{Debiasing Models}
\noindent \\
\textbf{Baseline.}
As the baseline, we used the large version of T5~\cite{2020t5}. T5 is an encoder-decoder transformer that is pre-trained on a cleaned Common Crawl collection, incorporating a mixture of supervised tasks through multi-task learning. To adopt T5-large for the debiasing task, we used WNC as the training dataset. Given our constraints with computational resources, we fine-tuned the model using LoRA adaptation~\cite{DBLP:conf/iclr/HuSWALWWC22}.

\begin{figure}[!t]
\centering
\footnotesize
\begin{tabular}{p{0.9\columnwidth}}
\toprule
\textbf{Debiasing Prompt} \\
\toprule
Transform the following biased sentence into an unbiased sentence from a news article by removing any subjective language or discriminatory undertones without changing its semantic meaning:
\\
\\
Biased Sentence:
\\
\\
\{\{sentence\}\}\\
\\
Unbiased Sentence: \\
\midrule
\textbf{Evaluator Prompt (shortened)} \\
\toprule
The input sentence from a news article is biased, it uses subjective
language or discriminatory undertones. The other sentence was debiased by a language model. Your task is to compare two sentences based on the following journalistic criteria. For each question in the checklist, select your response from \{1, 2, 3\}. \\
\\
The checklist is as follows:\\
1- Does the model produce unbiased text?\\
- If the text is unbiased, return 3\\
... \\
\\
Do not explain your decisions. \\
\\
Biased Sentence:\\
\\
\{\{sentence\}\}\\
\\
Model Output:\\
\\
\{\{model\_output\}\}\\
\\
Checklist Answers:\\
\bottomrule
\end{tabular}
\caption{Prompts for debiasing and evaluation. The full version of the evaluator prompt can be found at our source code.}
\label{tab:eval_prompts}
\end{figure}

\noindent
\textbf{Conversational LLMs.}
We evaluated the popular conversational LLMs: ChatGPT~\cite{ChatGPT} and GPT4~\cite{DBLP:journals/corr/abs-2303-08774} from OpenAI, and Llama2-70b-chat~\cite{touvron2023llama} which is an open source, popular alternative to ChatGPT. The models were adopted for conversational tasks using reinforcement learning with human feedback. In this way, individuals without technical expertise could easily interact with the models, making them suitable for integration into news organizations.

ChatGPT and Llama 2 are Autoregressive Language Models trained on large corpora from multiple sources from the web, with the objective of predicting the next word based on the preceding context. GPT4 is the advanced version of ChatGPT, capable of handling multi-modal input. While our task focused on textual input, we included GPT4 in our evaluations, because human evaluators preferred GPT4 outputs from various tasks over ChatGPT~\cite{DBLP:journals/corr/abs-2303-08774}. We used prompts, which are shown in Table~\ref{tab:eval_prompts}, for each of the conversational.

\section{Results} 

\begin{table}[!htp]
\adjustbox{width=0.95\columnwidth}{
    \centering
    \begin{tabular}{llllll}
    \toprule
    \textbf{ID} & \textbf{Grade} &\textbf{T5} & \textbf{Llama2}& \textbf{ChatGPT} & \textbf{GPT4}\\
    \midrule
    \multirow{ 3}{*}{\textbf{C1}} & \textbf{1} & 0.26 & 0.08 & 0.02 & 0\\
    & \textbf{2} & 0.40 & 0.06 & 0.14 & 0.38\\
    & \textbf{3} & 0.34 & \textbf{0.86} & 0.84 & 0.62 \\
    \midrule
    \multirow{ 3}{*}{\textbf{C2}} & \textbf{1} & 0.1 & 0.4 & 0.26 & 0.2 \\
    & \textbf{2} & 0.12 & 0.44 & 0.36 & 0.56\\
    & \textbf{3} & \textbf{0.78} & 0.16 & 0.38 & 0.24 \\
    \midrule
    \multirow{ 3}{*}{\textbf{C3}}  & \textbf{1} & 0.12 & 0.34 & 0.20 & 0.06\\
    & \textbf{2} & 0.06 & 0.4 & 0.48 & 0.68\\
    & \textbf{3} & \textbf{0.82} & 0.26 & 0.32 & 0.26 \\
    \midrule  
    \multirow{ 3}{*}{\textbf{C4}}  & \textbf{1} & 0.34 & 0.1 & 0.02 & 0 \\
    & \textbf{2} & 0.2 & 0 & 0 &  0.12\\
    & \textbf{3} & 0.46 & 0.9 & \textbf{0.98} & 0.88 \\
    \midrule
    \multirow{ 3}{*}{\textbf{C5}} & \textbf{1} & 0.14 & 0.44 & 0.42 & 0.42 \\
    & \textbf{2} & 0.08 & 0.46 & 0.56 & 0.5 \\
    & \textbf{3} & \textbf{0.78} & 0.1 & 0.02 & 0.08 \\
    \bottomrule
    \end{tabular}}
    \caption{The conversational LLMs are significantly better than the baseline at correcting bias and providing grammatically correct outputs (Student's T-test, p-value at 0.05), they have issues on preserving information, context and author's style.}
    \label{tab:human_eval}
\end{table}

\changemarker{Although BABE contains the biased text spans along with the labels, the dataset does not have the corrected versions of the biased texts. Therefore, we could not directly apply the evaluation criteria to the samples. For this reason, two} expert news editors from the team, as described in \S~\ref{sec:eval}, conducted the human evaluations voluntarily. \changemarker{Due to resource constraints, we split the models’ results into two parts for both evaluators. Each part contains the results from each model. One editor ranked the samples which they were responsible for, by using a 3-likert scale. During the ranking evaluation, the editor marked the samples they were unsure about, made notes and applied fact-checking to address the C2 and C3. The other editor reviewed the ranked samples while checking the notes, marked samples and the fact-checked ones. The editors regularly engaged in discussions to reach a consensus on disagreements and uncertain cases.} In total, we obtained 200 evaluations from the experts. Table~\ref{tab:human_eval} presents the frequency of ratings per criterion.
\\
\noindent \\
\textbf{RQ1: Debiasing Performance of the Conversational Models.} The conversational LLMs proved better than the baseline for debiasing. Surprisingly, Llama 2 demonstrated comparative results even though ChatGPT has been known to outperform others in various tasks~\cite{touvron2023llama}. The researchers of Llama 2's training regime - that the factual sources were prioritized in training samples - might account for its competitive performance in this task. The conversational LLMs also exhibited more grammatical correctness than the baseline. Nevertheless, some LLMs changed phrases they considered biased, while others removed words or sentiments that could be considered confrontational or impolite, but are not actually biased towards any particular group. For instance, GPT4 changed 'When carrying a firearm, you have the ultimate power of force in your control' to 'When carrying a firearm, you have a significant level of potential force at your disposal'.


The conversational LLMs performed worse than the baseline model in preserving information and context. These models introduced unnecessary amendments to the generated texts. In some cases, even created hallucinations. This issue is not unique to this study and has been reported in related studies, especially in the case of ChatGPT being used for various tasks~\cite{DBLP:journals/corr/abs-2302-04023}.Additionally, the news editors observed that Llama 2 introduced additional information not present in the input text, albeit factually accurate. For example, in a text mentioning 'Wilkens', the model replaced 'Wilkens' with 'Judge Wilkens'. The model may have memorized such information from its training dataset. This behavior by conversational models might harm the author's style. 

\begin{table}[!htp]
    \footnotesize
    \centering
    \begin{tabular}[width=\columnwidth]{llll}
    \toprule
    \textbf{ID} & \textbf{Llama2}& \textbf{ChatGPT} & \textbf{GPT4}\\
    \textbf{C1} & 0.0666 & -0.0489 & 0.1109 \\
    \textbf{C2} & -0.0145 & 0.0285 & 0.0018 \\
    \textbf{C3} & 0.1971 & 0.0280 & 0.0263 \\
    \textbf{C4} & 0.0597 & 0 & 0 \\
    \textbf{C5} & -0.0022 & 0.0454 & -0.0413 \\
    \bottomrule
    \end{tabular}
    \caption{The disagreement between the conversational tools as an evaluator and the expert evaluation is high, according to Cohen's Kappa.}
    \label{tab:bias_eval}
\end{table}

\noindent \\
\textbf{RQ2: Conversational LLMs as Evaluator:} We obtained rankings from the conversational LLMs and compared them with the expert rankings. As shown in Table~\ref{tab:bias_eval}, there is a high disagreement between the models and the expert evaluations. Additionally, we observed that the models rated the criteria, such as preserving factuality, grammar, with the highest score. In contrast, the ratings by the experts for these criteria were low. 

\section{Conclusion}
Through the editorial criteria, we showed that none of the conversational LLMs are perfect, even though they are good at debiasing. Specifically, they failed to preserve vital information and context, often leading to hallucinations. Employing these tools in a fully automatic editor can be dangerous, as they can create misinformation. 

Memorization also surfaces as an important aspect of LLM behavior. For this reason, to ensure a fair evaluation of debiasing tasks across news articles from different periods, Media bias researchers need to create benchmark datasets containing samples from time periods that is not covered within the training data of LLMs.

The assessments by the models are not close to those by the experts. \changemarker{We plan to increase the size of our annotations and the number of annotators to build a benchmark dataset for a fine-grained analysis of the models’ issues. We then} investigate advanced methods for automating the evaluation criteria and incorporating them to adapt the models.


\section*{Ethical Considerations and Limitations}
In this study, we assessed the efficiency of conversational LLMs in debiasing news articles, focusing solely on English samples from US Media. As a result, the generalizability of our conclusions to other languages and to media in other countries may be limited. 

The dataset employed in this research paper is derived from publicly accessible sources and is peer-reviewed. During the evaluation process, we refrained from disclosing the identities of the article publishers to the participating news editors, thereby preventing any potential influence on their evaluations.

\section*{Acknowledgement}
\changemarker{This research was partially funded by the JournalismAI Fellowship Programme 2024 of PolisLSE and vera.ai, which is co-financed by the European Union, Horizon Europe programme, Grant Agreement No 101070093. We also thank Kevin Nguyen, Saja Mortada, Khalid Waleed for their support.}

\section*{References}
\bibliographystyle{lrec_natbib}
\bibliography{lrec-coling2024-example}


\appendix




\end{document}